\documentclass[journal]{IEEEtran}
\usepackage{amsmath,amsfonts}
\usepackage{algorithmic}
\usepackage{algorithm}
\usepackage{array}
\usepackage[caption=false,font=normalsize,labelfont=sf,textfont=sf]{subfig}
\usepackage{textcomp}
\usepackage{stfloats}
\usepackage{url}
\usepackage{verbatim}
\usepackage{graphicx}
\usepackage{cite}
\usepackage{booktabs}
\usepackage{fontawesome}
\begin{document}

\title{SpaceSeg: A High-Precision Intelligent Perception Segmentation Method for Multi-Spacecraft On-Orbit Targets}

\author{Hao Liu, Pengyu Guo, Siyuan Yang, Zeqing Jiang, Qinglei Hu and Dongyu Li
\thanks{Hao Liu is with the Hangzhou International Innovation Institute, Beihang University, Hangzhou 311115, China (e-mail: lh@computer.org).}
\thanks{Pengyu Guo and Dongyu Li are with the School of Cyber Science and Technology, Beihang University, Beijing 100191, China (e-mail: 
 pengyuguo@buaa.edu.cn; dongyuli@buaa.edu.cn) (\emph{Corresponding author: Dongyu Li}).}
\thanks{Siyuan Yang is with the College of Computing and Data Science, Nanyang Technological University, Singapore 639798 (e-mail: siyuan.yang@ntu.edu.sg)}
\thanks{Zeqing Jiang is with the Ant Group, Shanghai 200122, China (e-mail: jiangzeqing.jzq@antgroup.com).}
\thanks{Qinglei Hu is with the School of Automation Science and Electrical Engineering, Beihang University, Beijing 100191, China (e-mail: huql$\_$buaa@buaa.edu.cn).}}



\maketitle

\begin{abstract}
With the continuous advancement of human exploration into deep space, intelligent perception and high-precision segmentation technology for on-orbit multi-spacecraft targets have become critical factors for ensuring the success of modern space missions. However, the complex deep space environment, diverse imaging conditions, and high variability in spacecraft morphology pose significant challenges to traditional segmentation methods. This paper proposes SpaceSeg, an innovative vision foundation model-based segmentation framework with four core technical innovations: First, the Multi-Scale Hierarchical Attention Refinement Decoder (MSHARD) achieves high-precision feature decoding through cross-resolution feature fusion via hierarchical attention. Second, the Multi-spacecraft Connected Component Analysis (MS-CCA) effectively resolves topological structure confusion in dense targets. Third, the Spatial Domain Adaptation Transform framework (SDAT) eliminates cross-domain disparities and resist spatial sensor perturbations through composite enhancement strategies. Finally, a custom Multi-Spacecraft Segmentation Task Loss Function is created to significantly improve segmentation robustness in deep space scenarios. To support algorithm validation, we construct the first multi-scale on-orbit multi-spacecraft semantic segmentation dataset SpaceES, which covers four types of spatial backgrounds and 17 typical spacecraft targets. In testing, SpaceSeg achieves state-of-the-art performance with 89.87$\%$ mIoU and 99.98$\%$ mAcc, surpassing existing best methods by 5.71 percentage points. The dataset and code are open-sourced at https://github.com/Akibaru/SpaceSeg to provide critical technical support for next-generation space situational awareness systems.
\end{abstract}


\section{INTRODUCTION}

\begin{figure}[!htb]
    \centering
    \includegraphics[width=0.98\linewidth]{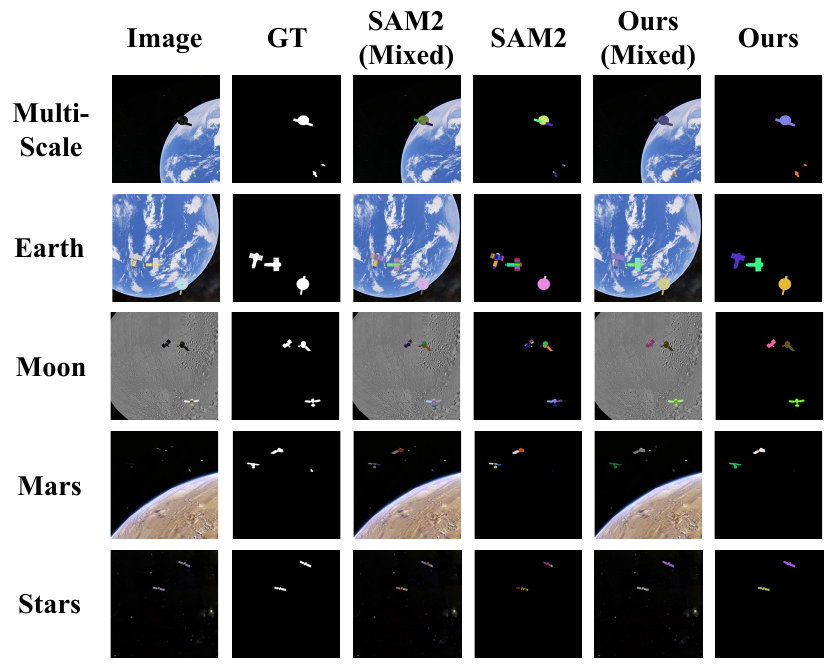}
    \caption{The target segmentation results of SpaceSeg. A comparison of target segmentation under different scales, spatial imaging backgrounds, and spacecraft categories on the selected dataset.} 
    \label{fig:tt}
\end{figure}

\IEEEPARstart{A}{s} humanity delves deeper into space exploration, intelligent perception technologies have emerged as pivotal elements ensuring the success of modern space missions. Tasks such as orbital debris removal, autonomous repair and refueling of deep-space probes, and threat detection and early warning for space targets demand precise identification and segmentation of key objects in spacecraft imagery. Such capabilities not only enhance mission intelligence but also play a critical role in safeguarding spacecraft operations and optimizing mission planning. However, the task of spacecraft image segmentation faces unique challenges, including the complexity of image acquisition, diverse imaging conditions, and high variability in spacecraft morphology.

Recent years have witnessed remarkable advancements in image segmentation techniques driven by deep learning, particularly in terrestrial and conventional imaging scenarios. Image segmentation, a fundamental task in computer vision, underpins various applications in visual understanding. By partitioning images into semantically meaningful regions, segmentation techniques have enabled downstream tasks across fields such as optical remote sensing image segmentation\cite{Zhou2024MeSAM:,DIAKOGIANNIS202094} and medical image segmentation\cite{Chen2024SAM2-Adapter:,Ma2024,Huang_2024,xiong2024sam2unetsegment2makes}. While specialized architectures have achieved state-of-the-art performance in these domains, directly applying such methods to spacecraft scenarios presents several obstacles. The challenges include limited data availability with high annotation costs, as well as unique space-specific issues such as strong light reflections, shadow occlusions, and complex deep-space backgrounds. These factors significantly hinder the generalizability of traditional segmentation methods, making them insufficient for real-world space applications.

Moreover, with the large-scale deployment of satellite constellations and the rapid development of commercial space activities, the number of spacecraft in orbital space is growing exponentially. This is particularly evident in the Low Earth Orbit (LEO) region, where the overlapping deployment of multiple commercial satellite constellations has resulted in a densely populated spatial distribution of spacecraft. This increases the risk of collisions and significantly complicates the management of orbital resources. Consequently, achieving accurate identification, positioning, and status monitoring of multiple spacecraft has become increasingly critical. Traditional target recognition methods based on manual feature extraction can no longer meet the demands of the current space environment, characterized by multiple targets, high density, and rapid changes. This drives the need to explore a new generation of multi-spacecraft target segmentation methods, aiming to provide more precise and real-time space situational awareness. Such advancements can offer essential technological support for tasks such as space debris monitoring, spacecraft safety operations, orbital traffic management, and non-cooperative space target recognition.

To address these challenges, this study introduces SpaceSeg, a novel vision-based segmentation model specifically designed for multi-object and multi-spatial backgrounds in spacecraft imagery. SpaceSeg aims to overcome the limitations of existing techniques by leveraging the transformative potential of Vision Foundation Models (VFMs) \cite{kirillov2023segment,ravi2024sam2segmentimages,li2024omgsegmodelgoodsegmentation}. A prominent example is the Segment Anything Model (SAM) \cite{kirillov2023segment} and its successor SAM2 \cite{ravi2024sam2segmentimages}, which has been pre-trained on larger datasets and refined with architectural improvements. SAM2, with its enhanced generalization capability, serves as an ideal foundation for tackling the semantic understanding required in complex space environments.

Building upon SAM2, this research proposes an improved backbone network architecture tailored for space perception tasks. we froze the hierarchical image encoder pre-trained with MAE, endowing it with higher spatial resolution in multi-scale feature extraction. Without providing additional prompts, our unique domain-specific knowledge for spacecraft data was injected into the original SAM2 mask decoder. To generalize the model under limited labeled data conditions, SpaceSeg employs spatially selective dual-stream adaptive fine-tuning to train task-specific flows between the mask decoder and prompt decoder. This maximizes the utilization of knowledge from spacecraft samples, enabling precise semantic feature comprehension. Building on this, we further introduced the Multi-Scale Hierarchical Attention Refinement Decoder (MSHARD), which extends traditional decoding structures by incorporating key steps such as multi-scale feature hierarchy, hierarchical attention refinement and cross-level feature integration guided by attention. This design captures multi-scale features and global spatiotemporal correlations in spacecraft imagery. For multi-spacecraft scenarios, we adopted Connected Component Analysis (CCA) to independently label each target, enabling parallel and efficient instance differentiation and labeling. To address the challenges of limited data and varying imaging characteristics in deep space environments, we devised a Spatial Domain Adaptive Transform (SDAT) framework. This framework integrates various geometric transformations and noise simulation methods, ensuring stable perceptual capabilities in highly dynamic, low signal-to-noise ratio deep space environments. To accurately evaluate the performance of spacecraft multi-object segmentation, we incorporated binary cross-entropy loss, IoU prediction loss, and robust design into the loss function. This ensures high sensitivity to details and boundaries in stringent aerospace scenarios while maintaining the consistency of results. Furthermore, addressing the unique demands of aerospace missions, this study designs a data-engine-driven space image segmentation dataset. These innovations comprehensively evaluate model performance across multi-scale and multi-granularity scenarios, significantly boosting the model's robustness and adaptability to the complexities of space environments.
Our contributions are summarized as follows:

\begin{itemize}
    \item We propose SpaceSeg, an innovative on-orbit multi-spacecraft target segmentation model. Compared with other mainstream advanced segmentation methods, the proposed approach introduces a comprehensive solution for multi-target and multi-scale tasks, demonstrating significant advantages in achieving high precision in segmentation performance.
    \item To support this work, we have developed and released a publicly available on-orbit multi-spacecraft multi-scale target segmentation dataset. This dataset not only facilitates spacecraft target segmentation but also serves as a valuable resource for advancing research on other space vision tasks, especially in complex deep-space environments.
    \item Unlike traditional manually designed segmentation networks, this work is the first to leverage and improve vision foundation models, enabling the acquisition of powerful prior background knowledge. By integrating robust feature extraction and fusion methods, the proposed model achieves optimal performance in space target segmentation tasks. Furthermore, it demonstrates how to leverage the strong generalization capabilities and transferability of visual foundation models to a broader range of task domains, underscoring its potential for applications beyond the immediate scope of this study.
\end{itemize}

\section{RELATED WORKS}
\subsection{Off-Earth Imaging and On-Orbit Spacecraft Perception}
Off-Earth imaging refers to the technologies and methods for imaging and observing celestial bodies or space beyond Earth. In the field of on-orbit spacecraft perception, it is also known as satellite-to-satellite imaging or space-to-space (S2S) imaging, involving the use of space-based sensors to capture high-resolution images of resident space objects (RSOs). These objects can include active or inactive spacecraft, rocket bodies, or uncontrolled spacecraft debris.

Recent research on on-orbit spacecraft perception has emphasized enhancing monitoring, analysis, and decision-making in the space environment. A key focus in this field is on improving imaging capabilities and augmenting autonomous decision-making processes. For instance, \cite{Li2017Image} introduced an image quality enhancement method for on-orbit remote sensing cameras using the invariable modulation transfer function, which significantly improves imaging resolution. Similarly, \cite{felicatti2023hysim} developed a tool for space-to-space hyperspectral imaging, enabling enhanced target detection in hyperspectrally-resolved imagery. To address spacecraft target recognition challenges, \cite{9602108} proposed a novel method based on ellipse detection, which has shown promise in improving detection accuracy. Recent advances in artificial intelligence have also driven progress, with techniques such as YOLOv2 applied to spacecraft object detection \cite{8600520}. Additionally, \cite{8778767} leveraged the artificial bee colony algorithm with multipeak optimization to detect non-cooperative spacecraft targets. Despite these advancements, the field remains in its early stages, warranting further investigation and development.

\subsection{Image Segmentation and Spacecraft Target Segmentation}
Image segmentation is a fundamental task in computer vision, widely studied due to its extensive applicability in fields such as medical imaging, autonomous driving, and remote sensing. Traditional methods\cite{937505,10.1145/1015706.1015720} heavily rely on handcrafted features and lack robustness against complex visual variations. With the advent of deep learning, architectures like U-Net\cite{DBLP:journals/corr/RonnebergerFB15}, DeepLab\cite{DBLP:journals/corr/ChenPK0Y16}, and Mask R-CNN\cite{DBLP:journals/corr/HeGDG17} have significantly advanced segmentation accuracy and scalability. U-Net effectively integrates multi-scale features through skip connections, becoming a benchmark in medical image segmentation. The DeepLab series expands the receptive field using dilated convolutions and introduces the Atrous Spatial Pyramid Pooling (ASPP) module to capture multi-scale contextual information, markedly improving segmentation precision.

Spacecraft target segmentation presents unique technical challenges and extends beyond traditional image segmentation tasks. One primary issue lies in the highly reflective materials and intricate geometries of spacecraft surfaces, which often cause conventional segmentation models to over-segment or under-segment. Additionally, the distinct conditions of the space environment introduce significant hurdles when applying deep learning techniques to in-orbit spacecraft. These challenges include the scarcity of labeled data and the need for high-precision models capable of operating efficiently in constrained environments. To address these issues, \cite{s22114222} developed a spacecraft target segmentation model leveraging DeepLabv3+ integrated with dilated convolution to enhance segmentation accuracy. Similarly, \cite{9465153} proposed a mean shift clustering algorithm to detect hypervelocity impact damage on spacecraft surfaces. Despite these advancements, research in this field remains in an exploratory phase, requiring further investigation and innovation.

\subsection{Spacecraft Datasets}
Obtaining real in-orbit spacecraft images is both rare and sensitive due to the classified nature of such data. Consequently, the majority of spacecraft image datasets are generated through simulation. Current research on spacecraft segmentation datasets primarily focuses on component-level segmentation and ground-based scene testing. For example, \cite{10115564} synthesized a prototype dataset comprising 2D monocular images of unmanned spacecraft, while \cite{10380525} developed a dataset tailored for the detection and segmentation of key spacecraft components. Expanding on this, \cite{hoang2021spacecraftdatasetdetectionsegmentation} introduced a comprehensive spacecraft dataset designed for detection, segmentation, and component recognition tasks. The existing datasets primarily focus on single spacecraft and are primarily tested in ground-based scenarios, lacking data for multi-spacecraft cooperative perception, diverse spatial backgrounds, and multi-scale characteristics. As a result, they fail to adequately simulate the challenges of real space environments and cannot meet the demands of on-orbit services. Therefore, to advance on-orbit multi-spacecraft perception, it is essential to design datasets that incorporate multi-spatial backgrounds and multi-scale characteristics, addressing the diverse challenges of space environments.

\subsection{Vision Foundation Models}
In recent years, the development of vision foundation models, pre-trained models, and large models has become a research hotspot in the field of computer vision. Vision foundation models provide general feature representations for various vision tasks through unified architecture design and task generalization capabilities. Pre-trained models, by leveraging unsupervised or self-supervised learning on large-scale datasets, capture rich semantic information and contextual features, providing powerful initialization weights for downstream tasks, which significantly improve model performance and training efficiency. SAM and SAM2 models are examples of vision foundation models constructed using pre-training. On this basis, the emergence of large models has further pushed the boundaries of capabilities in vision tasks. By enlarging model size, incorporating multi-modal learning, and optimizing training strategies, these models exhibit wide-ranging applicability across tasks, domains, and even modalities. For example, models based on the Transformer architecture, such as Vision Transformer (ViT)\cite{DBLP:journals/corr/abs-2010-11929} and CLIP\cite{DBLP:journals/corr/abs-2103-00020}, have achieved significant progress in tasks like image classification, object detection, and image-text matching. At the same time, they have sparked widespread discussions on parameter scale, data quality, and computational resource requirements. Currently, the application of SAM series models in the medical\cite{Chen2024SAM2-Adapter:,Ma2024,Huang_2024,xiong2024sam2unetsegment2makes} and remote sensing\cite{Zhou2024MeSAM:,DIAKOGIANNIS202094} domains is relatively abundant. However, the field of spacecraft intelligent perception still primarily relies on artificial neural network design, and applying vision foundation models and pre-trained models in this area remains a research gap requiring further exploration.

\section{METHODOLOGY}

\begin{figure*}[ht]
    \centering
    \includegraphics[width=0.98\linewidth]{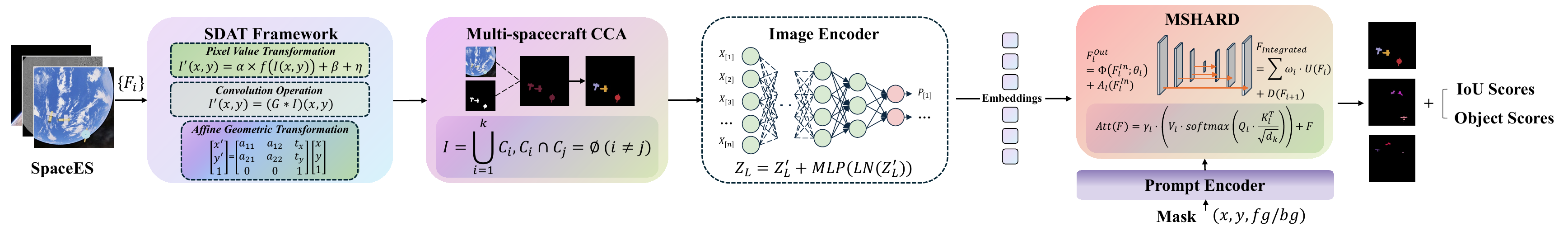}
    \caption{The SpaceSeg architecture. Spatial imaging images are processed through the SDAT Framework for strong robustness and generalization. The results are then passed through the Multi-spacecraft CCA mechanism to determine multi-spacecraft task recognition outcomes, which are subsequently fed into the Image Encoder and then into MSHARD. On the other hand, the Mask, by sampling, becomes the prompt and is fed into MSHARD via the Prompt Encoder. Finally, on-orbit multi-spacecraft object segmentation results are obtained.} 
    \label{fig:pipe}
\end{figure*}

This section aims to systematically elaborate on the design and optimization of the proposed SpaceSeg model for the multi-object segmentation tasks of spacecraft. The architecture diagram of SpaceSeg is shown in Figure \ref{fig:pipe}. Overall, this framework leverages and extends the image encoder and mask decoder of the large-scale pre-trained model (SAM2). On one hand, it inherits the general visual features learned from large-scale and diverse datasets. On the other hand, it effectively enhances the recognition and segmentation capabilities for spacecraft imagery in deep space environments through domain adaptation for spatial perception and a multi-scale attention mechanism. By introducing the Multi-Scale Hierarchical Attention Refinement Decoder, Multi-spacecraft Connected Component Analysis, Spatial Domain Adaptation Transform Framework, and a specialized Loss Function Design, the model enhances feature extraction capabilities and achieves higher accuracy in multi-object, multi-scale segmentation tasks for on-orbit spacecraft images. Through this series of multi-level and multi-perspective enhancements, the SpaceSeg framework demonstrates exceptional segmentation accuracy and adaptability in spacecraft segmentation tasks.

\subsection{Space Perception Backbone and Spatial Selective Dual-Stream Adaptive Fine-tuning}
Our SpaceSeg framework is fundamentally built upon the robust image encoder and mask decoder components of the SAM2 model. Specifically, we employ the MAE pre-trained Hiera\cite{ryali2023hierahierarchicalvisiontransformer} image encoder within SAM2 and freeze its weights to preserve the rich visual representations learned from pre-training on large-scale datasets. Compared to the standard ViT encoder used in SAM1, Hiera adopts a hierarchical structure that allows for multi-scale feature capture, making it more suitable for extracting spatial multi-scale information. Specifically, given an input image  \( I \in \mathbb{R}^{3 \times H \times W} \),  where  \(H\) denotes height and  \(W\) denotes width, Hiera will output four hierarchical features \( X_i \in \mathbb{R}^{C_i \times \frac{H}{2^{i+1}} \times \frac{W}{2^{i+1}}} (i \in \{1, 2, 3, 4\})\). For Hiera-S, \( C_i \in \{96, 192, 384, 768\}. \)

Furthermore, we utilize the mask decoder module from the original SAM2 model, initializing its weights with the pre-trained SAM2 parameters, and then perform spatial perception domain-specific adaptation during the training of our adapter. Notably, we do not provide any additional prompts as input to the original SAM2 mask decoder.

Subsequently, we learn and inject task-specific spatial knowledge \(F_i\) from the dataset into the network via spatial selective dual-stream adaptive fine-tuning, which selectively trains the mask decoder and prompt decoder through dual streams to achieve knowledge transfer of spacecraft samples. We adopt the concept of prompting, sampling the input mask to obtain the prompt, which leverages the fact that foundational models like SAM2 have been trained on large-scale datasets. By introducing task-specific knowledge through appropriate prompts, we enhance the model's generalization capabilities on downstream tasks, especially in scenarios where annotated data is scarce.

\subsection{Multi-Scale Hierarchical Attention Refinement Decoder}
In this section, we introduce a novel architecture specifically designed for spacecraft segmentation tasks—the Multi-Scale Hierarchical Attention Refinement Decoder (MSHARD). Built on the SAM2 framework, this approach incorporates hierarchical attention mechanisms and multi-scale feature refinement to address unique challenges in spatial imagery, including variable lighting conditions, diversity and scale variations, and the complex geometry of spacecraft. Given an input image \( I \in \mathbb{R}^{H \times W \times 3} \) containing spacecraft targets and a set of prompt points \( P = \{(x_i, y_i)\}_{i=1}^n \), our goal is to generate a precise segmentation mask \( M \in \{0, 1\}^{H \times W} \).

\begin{figure*}[ht]
    \centering
    \includegraphics[width=0.98\linewidth]{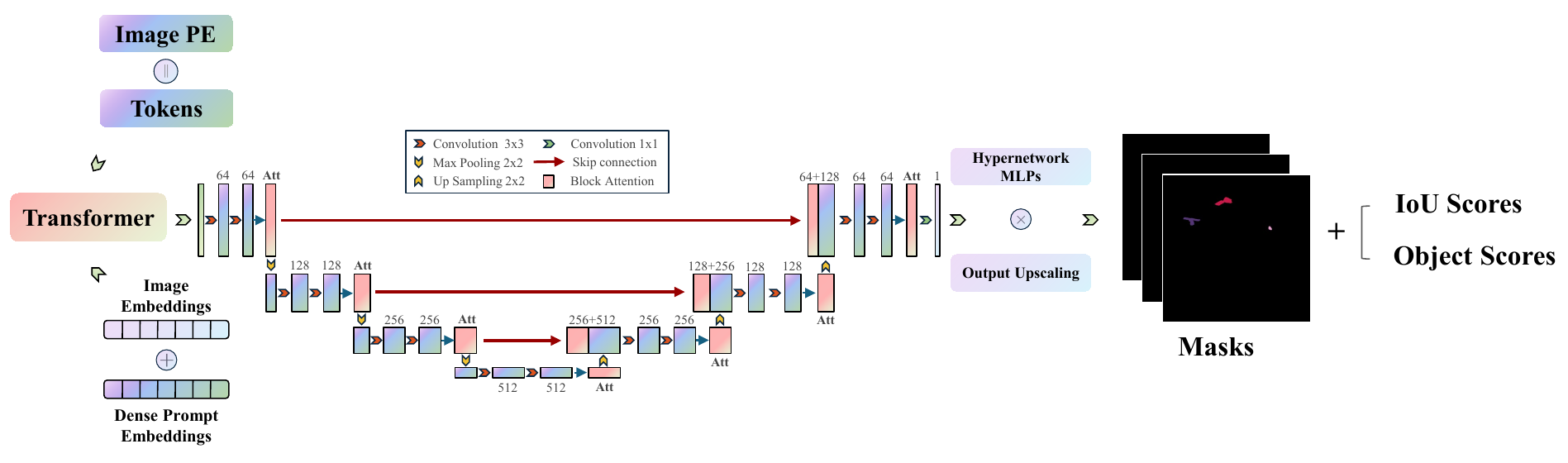}
    \caption{The MSHARD architecture employs a hybrid fusion strategy for multimodal feature integration. Specifically, image embeddings undergo element-wise summation with dense vector representations of mask prompt by sampling, whereas positional encodings of visual elements are concatenated with semantic token sequences before joint processing through transformer layers. The hierarchical feature aggregation is achieved through a U-Net-style architecture enhanced with level-specific attention mechanisms to capture spatial domain detailed features. This core computational framework interfaces with parallel output heads: the mask reconstruction is refined through upsampling operators, while geometry quality evaluation (IoU) and object semantic confidence metrics are synchronously output via hypernetwork-based MLP branches.} 
    \label{fig:MSHARD}
\end{figure*}

The proposed MSHARD introduces several key innovations based on a Transformer-based architecture. It consists of multiple hierarchical stages, each operating at different spatial resolutions. Unlike the traditional SAM 2 decoder, our architecture incorporates three key modifications: (1) a multi-scale feature hierarchy, (2) hierarchical attention refinement, and (3) cross-level feature integration guided by attention. MSHARD retains a U-Net-like structural style while embedding attention mechanisms at each level to better capture fine-grained spacecraft details and global contextual information. Figure \ref{fig:MSHARD} illustrates the architecture of MSHARD. 

We adopted a hierarchical U-Net structure with four resolution levels \(\{l_1, l_2, l_3, l_4\}\) and the level-specific attention mechanism, where the resolution at each level li is given by \(R_i = R_0 / 2^i\), as shown in Figure \ref{fig:MSHARD}. The feature maps at each level undergo the following processing:
\begin{equation}
    F_l^{\text{out}} = \Phi(F_l^{\text{in}}; \theta_l) + A_l(F_l^{\text{in}})
\end{equation}
where \(\Phi(\cdot)\) represents the feature refinement operation, and \(A_l(\cdot)\) denotes the level-specific attention mechanism. The feature refinement operation incorporates parallel convolution branches with different kernel sizes to effectively capture multi-scale information. MSHARD's hierarchical attention mechanism operates across multiple scales, allowing the model to focus on relevant spacecraft features while suppressing background noise. 

For each level \(l\), the attention is computed as:
\begin{equation}
    A_l(F) = \gamma_l \cdot \left(V_l \cdot \text{softmax}\left(Q_l \cdot K_l^T / \sqrt{d_k}\right)\right) + F
\end{equation}
where \(Q_l\),\(K_l\),\(V_l\) are learnable query, key, and value transformations; \(\gamma_l\) is a learnable scale parameter; and \(d_k\) is the dimension of the key vectors. This method can adaptively focus on relevant features at each scale, thereby capturing long-range dependencies in spacecraft structures while preserving fine details for precise segmentation.

For the cross-scale feature integration mechanism, to enhance information flow between different resolution levels, we incorporate hierarchical attention in the design:
\begin{equation}
    F_{\text{integrated}} = \sum \omega_i \cdot U(F_i) + D(F_{i+1})
\end{equation}
where \(U(\cdot)\) represents the upsampling operation, \(D(\cdot)\) denotes the downsampling operation, and \(\omega_i\) are learnable weight parameters.

\subsection{Multi-spacecraft Connected Component Analysis}
In the multi-spacecraft target segmentation task, we employ Connected Component Analysis (CCA) to independently identify and label target instances. This method is based on the concept of connectivity in graph theory, treating adjacent sets of foreground pixels in a binary image as distinct connected components.Let the binary image be \(I(x,y)\), where the foreground pixel value is 1, and the background pixel value is 0. For any two pixels \(p_1(x_1,y_1)\) and \(p_2(x_2,y_2)\), if there exists a path P:

\begin{equation}
P = \{p_i \mid i = 1,2,\dots,n\}
\end{equation}

Such that:

\begin{align*}
    & 1. \; p_1 \text{ and } p_n \text{ are the start and end points, respectively;} \\
    & 2. \; \forall i \in [1, n], \; I(p_i) = 1; \\
    & 3. \; \forall i \in [1, n-1], \; p_i \text{ and } p_{i+1} \text{ are adjacent in the } \\
    & \quad \text{ 8-neighborhood. }
\end{align*}

Then \(p_1\) and \(p_2\) are said to be connected. Based on this definition, the image can be divided into \(k\) disjoint connected components:

\begin{equation}
I = \bigcup_{i=1}^k C_i, \quad C_i \cap C_j = \emptyset \quad (i \neq j)
\end{equation}

Each connected component \(C_i\) represents an independent spacecraft instance. By assigning a unique label to each connected component, we obtain a labeled image \(L(x,y)\):

\begin{equation}
L(x,y) = i \quad \text{if} \quad (x,y) \in C_i
\end{equation}

Due to the possibility of overlap or occlusion among multiple spacecraft targets, a secondary analysis is triggered for suspicious connected components (e.g., those exhibiting abnormal shapes or large discrepancies within the mask). Graph Cut is then employed to further refine the segmentation. If a connected component is split into multiple parts, each part is assigned a new target ID. However, if it is determined to be the same target occluding itself, the original ID is retained.

\begin{figure}[b]
    \centering
    \includegraphics[width=0.98\linewidth]{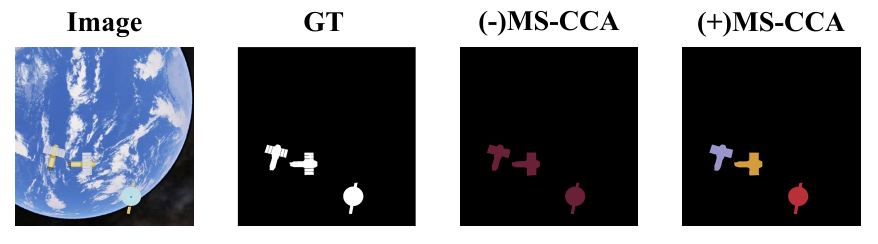}
    \caption{The impact comparison of multi-spacecraft connected component analysis on actual segmentation effectiveness. It can be clearly observed that the use of the MS-CCA method achieves accurate target segmentation for multiple spacecraft.} 
    \label{fig:CCA}
\end{figure}

This connectivity-based instance segmentation method effectively handles multi-target scenarios in spacecraft images, laying the groundwork for subsequent instance-level feature extraction and trajectory analysis. In practical implementation, we employ a two-pass scanning algorithm based on equivalence classes, with a time complexity of \(O(N)\), where \(N\) is the total number of pixels in the image. Using this method, the model can accurately identify and segment multiple spacecraft in multi-spacecraft segmentation scenarios, as shown in Figure \ref{fig:CCA}.

\subsection{Spatial Domain Adaptation Transform Framework}

\begin{figure}[h]
    \centering
    \includegraphics[width=0.98\linewidth]{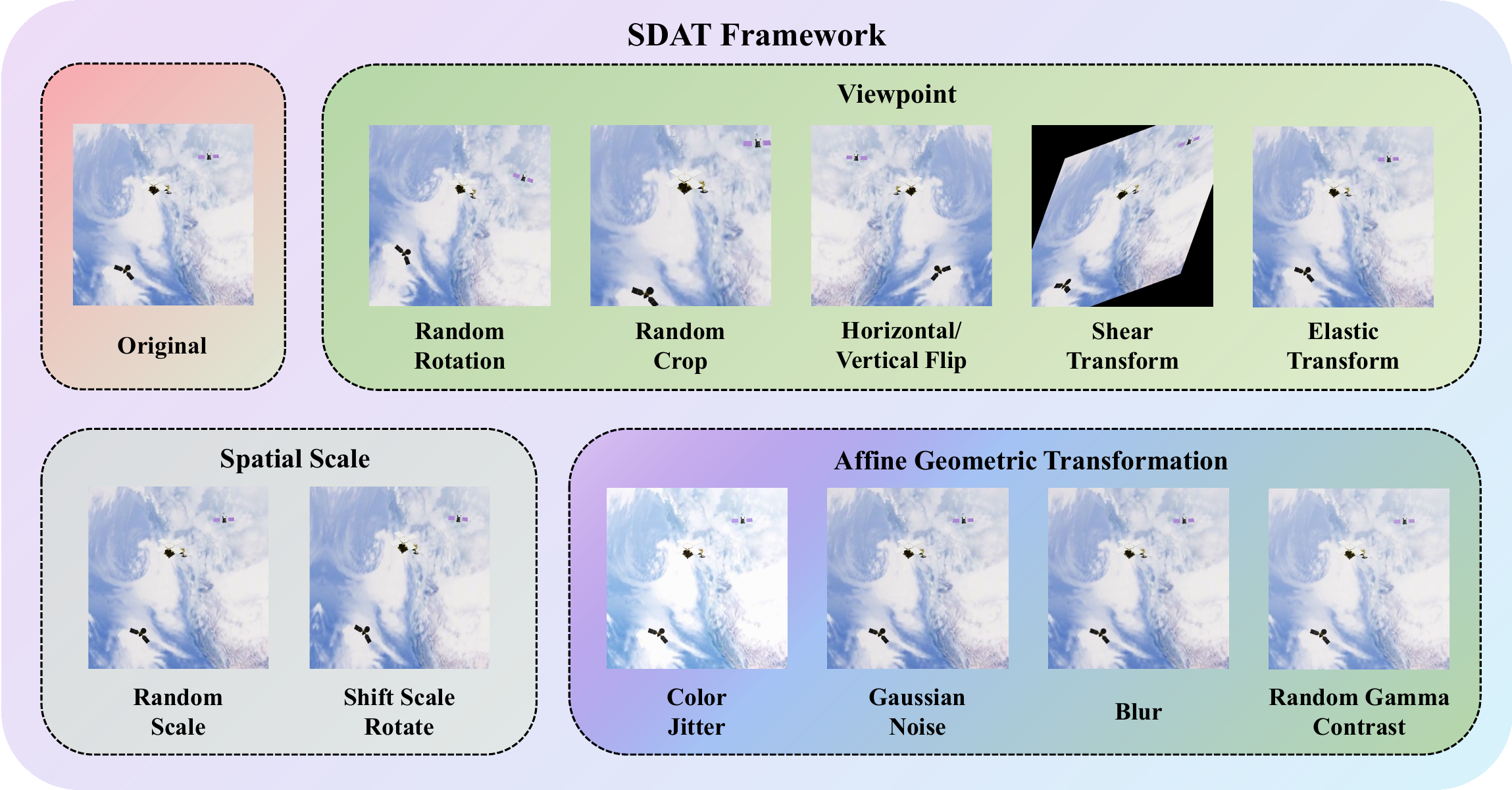}
    \caption{The structural composition and transformation effects of the SDAT framework. Adopting a combination of multiple strategies enables the model to achieve strong robustness and generalization for imaging and target segmentation in space missions.} 
    \label{fig:SDAT}
\end{figure}

In spacecraft spatial perception and imaging tasks, the availability of training data resources and the imaging characteristics of deep-space environments significantly affect the specific outcomes of image perception. Insufficient data resources for spacecraft spatial perception hinder visual models from effectively and fully learning imaging features, leading to reduced generalization ability and diminished detail recognition. Moreover, the imaging characteristics of complex deep-space environments—such as high dynamics and low signal-to-noise ratios—also impact the stability of the model's perceptual capabilities.
To address the issues of limited training data for spatial visual models and the need for strong robustness in perceptual features, we have designed the Spatial Domain Adaptation Transform Framework (SDAT Framework) based on the characteristics of spatial perception tasks. This framework primarily consists of a series of spatial zero-shot adaptation transformation strategies, divided into three sub-strategy segments: viewpoint, spatial scale, and perception environment, as illustrated in Figure \ref{fig:SDAT}.

In the SDAT framework, we propose a unified mathematical framework for spatial domain adaptive transformation. All transformations can be categorized into three basic operations: affine geometric transformation, pixel value transformation, and convolution operation.

For affine geometric transformation, we adopt a unified matrix representation:
\begin{equation}
\begin{bmatrix} 
x' \\ 
y' \\ 
1
\end{bmatrix} = 
\begin{bmatrix}
a_{11} & a_{12} & t_x \\
a_{21} & a_{22} & t_y \\
0 & 0 & 1
\end{bmatrix}
\begin{bmatrix}
x \\
y \\
1
\end{bmatrix}
\end{equation}

This affine transformation unifies the following transformations: random rotation \eqref{v1}, random crop \eqref{v2},  horizontal/vertical flip \eqref{v3}\eqref{v4}, shear transform \eqref{v5}, random scale \eqref{v7} and shift scale rotate \eqref{v8}.

For pixel value transformation, we propose a unified form:
\begin{equation}
I'(x,y) = \alpha \times f(I(x,y)) + \beta + \eta
\end{equation}

This formula unifies the following transformations: color jitter \eqref{v9}, Gaussian noise \eqref{v10} and random gamma contrast \eqref{v12}.

For operations involving spatial domain convolution, we present a unified representation:
\begin{equation}
I'(x,y) = (G * I)(x,y)
\end{equation}

This includes blur \eqref{v11} and elastic transform \eqref{v6}.

Based on this mathematical framework, SDAT implements eleven specific image enhancement strategies, each being a specialized implementation of one or more basic operations from the framework. This unified mathematical description not only provides excellent scalability but also establishes a theoretical foundation for combining different transformations, as summarized in the following paragraphs.

\subsubsection{Random Rotation} In the space environment, a spacecraft's attitude and orientation frequently change, causing imaging devices to capture the target at various angles. Randomly rotating images aids the perception model in learning to recognize object features under different attitudes and rotation angles, enhancing adaptability to variations in spacecraft posture. The method can be operated by \eqref{v1} where \( \theta \in [-180^\circ, 180^\circ] \).

\begin{equation}
\label{v1}
\begin{bmatrix} 
x' \\ y'
\end{bmatrix} = 
\begin{bmatrix}
\cos\theta & -\sin\theta \\
\sin\theta & \cos\theta
\end{bmatrix}
\begin{bmatrix}
x \\ y
\end{bmatrix}
\end{equation}

\subsubsection{Random Crop} As spacecraft move through space, they may only capture partial regions of a target spacecraft. Randomly cropping images forces the model to focus on local features, improving its recognition capability when information is missing, such as handling partial occlusions or limited fields of view in real-world scenarios. The method can be operated by \eqref{v2} where \( x \in [0, W-w], y \in [0, H-h] \) and \(W,H \) are original dimensions.

\begin{equation}
\label{v2}
C(I) = I[x:x+h, y:y+w]
\end{equation}

\subsubsection{Horizontal/Vertical Flip} In space, objects can be observed from any direction, lacking a fixed up, down, left, or right. Flipping images horizontally or vertically simulates the diverse directional inputs a spacecraft may encounter, helping the model learn image features under directional changes and enhancing its ability to recognize both symmetric and asymmetric objects. The method can be operated by \eqref{v3} or \eqref{v4} where \(W,H\) are image dimensions.

\begin{equation}
\label{v3}
F_h(x,y) = (W-x, y)
\end{equation}

\begin{equation}
\label{v4}
F_v(x,y) = (x, H-y)
\end{equation}

\subsubsection{Shear Transform} Shear transformations simulate the tilt of perception elements or the oblique viewing effects of objects in space. During spacecraft missions, imaging devices may observe targets at non-orthogonal angles, resulting in images with shear-induced distortions. By applying shear transformations, the model can learn and adapt to these angular variations, improving recognition accuracy from different observation perspectives. The method can be operated by \eqref{v5}.

\begin{equation}
\label{v5}
\begin{bmatrix} 
x' \\ y'
\end{bmatrix} = 
\begin{bmatrix}
1 & s_x \\
s_y & 1
\end{bmatrix}
\begin{bmatrix}
x \\ y
\end{bmatrix}
\end{equation}

\subsubsection{Elastic Transform} In space environments, images may be affected by special factors such as gravitational field fluctuations or magnetic interference, leading to local deformations in the captured images. Elastic transformations apply localized elastic distortions to simulate these deformation interferences. This strategy enhances the model's sensitivity to minor deformations, boosting recognition capabilities under complex space conditions. The method can be operated by \eqref{v6} where \(\delta x, \delta y = \alpha \times (G_\sigma * \Delta)\) and \(G_\sigma\) is Gaussian filter with standard deviation \(\sigma\).

\begin{equation}
\label{v6}
E(x,y) = (x + \delta x(x,y), y + \delta y(x,y))
\end{equation}

\subsubsection{Random Scale} The constantly changing distance to a target spacecraft results in varying object scales within images. Random scaling simulates observation effects at different distances by altering image sizes, helping the model learn and adapt to features of the observed spacecraft across various scales, thereby improving perception accuracy at different distances. The method can be operated by \eqref{v7}.

\begin{equation}
\label{v7}
\begin{bmatrix} 
x' \\ y'
\end{bmatrix} = 
\begin{bmatrix}
s_x & 0 \\
0 & s_y
\end{bmatrix}
\begin{bmatrix}
x \\ y
\end{bmatrix}
\end{equation}

\subsubsection{Shift Scale Rotate} This strategy combines translation, scaling, and rotation transformations to comprehensively simulate changes in a spacecraft's position and attitude in space. Translating the image simulates positional changes; scaling simulates distance variations; rotating simulates attitude adjustments. This combined transformation enables the model to learn how to accurately perceive and recognize image content under dynamic spatial conditions.The method can be operated by \eqref{v8}.

\begin{equation}
\label{v8}
\begin{bmatrix} 
x' \\ y' \\ 1
\end{bmatrix} = 
\begin{bmatrix}
s\cos\theta & -s\sin\theta & t_x \\
s\sin\theta & s\cos\theta & t_y \\
0 & 0 & 1
\end{bmatrix}
\begin{bmatrix}
x \\ y \\ 1
\end{bmatrix}
\end{equation}

\subsubsection{Color Jitter} Space presents extremely complex lighting conditions, such as intense sunlight, Earth's reflected light, and the darkness of space. Color jitter adjusts image parameters like brightness, contrast, saturation, and hue to simulate different cosmic illumination scenarios, assisting the model in maintaining stable performance and adaptability across various lighting environments.The method can be operated by \eqref{v9}.

\begin{equation}
\label{v9}
J(I) = (\beta \cdot I + b) \times c
\end{equation}

\subsubsection{Gaussian Noise} Addition In space, sensors may be affected by cosmic rays, solar wind, and other interferences, introducing noise into images. Adding Gaussian noise simulates sensor noise in the space environment, enhancing the model's robustness in noisy conditions and enabling it to accurately extract valuable information even under low signal-to-noise ratios.The method can be operated by \eqref{v10} where \(\eta \sim \mathcal{N}(0, \sigma^2)\).

\begin{equation}
\label{v10}
N(I) = I + \eta
\end{equation}

\subsubsection{Blur} Due to spacecraft motion or limitations of imaging equipment, captured images may exhibit blurring caused by optical errors or sensor distortions. Applying blur processing simulates situations like spacecraft motion blur or defocusing, improving the model's ability to recognize key features in blurred images during actual spacecraft perception tasks. The method can be operated by \eqref{v11} where \(G_\sigma = \frac{1}{2\pi\sigma^2}e^{-\frac{x^2+y^2}{2\sigma^2}}\).

\begin{equation}
\label{v11}
B(I) = G_\sigma * I
\end{equation}

\subsubsection{Random Gamma Contrast} Dramatic lighting variations in space may cause images to be overly bright or dark. Adjusting the gamma value of images simulates imaging effects under different illumination conditions. This strategy allows the model to adapt to high dynamic range lighting changes, enhancing robustness in cosmic imaging environments. The method can be operated by \eqref{v12} where \( \gamma \in [0.8, 1.2] \).

\begin{equation}
\label{v12}
G(I) = I^\gamma
\end{equation}

\subsection{Loss Function}
We designed a loss function tailored for multi-spacecraft segmentation tasks. The training objective integrates multiple components: \(L_{\text{seg}} = -\sum \left( y_i \log(p_i) + (1-y_i) \log(1-p_i) \right)\) is used to measure segmentation performance, enhances sensitivity to spacecraft edges and detailed features, making it suitable for handling fine spacecraft structures and distinguishing closely spaced spacecraft; \(L_{\text{IoU}}\) evaluates the accuracy of mask quality prediction,  ensures reliability, aiding in identifying uncertain segmentation results caused by lighting or posture changes and evaluating segmentation quality in complex space backgrounds; \(L_{\text{stability}}\) ensures the stability of predictions across different scales and improves model stability in harsh space environments, preventing training crashes under extreme conditions and addressing noise issues in space environments. The parameters \(\lambda_1\) and \(\lambda_2\) are balancing hyperparameters. The total loss is given by Equation \ref{total_loss}.

\begin{equation}
\label{total_loss}
L_{\text{total}} = L_{\text{seg}} + \lambda_1 L_{\text{IoU}} + \lambda_2 L_{\text{stability}}
\end{equation}

\section{EXPERIMENTS}
\subsection{Datasets}

\begin{figure}[t]
    \centering
    \includegraphics[width=0.98\linewidth]{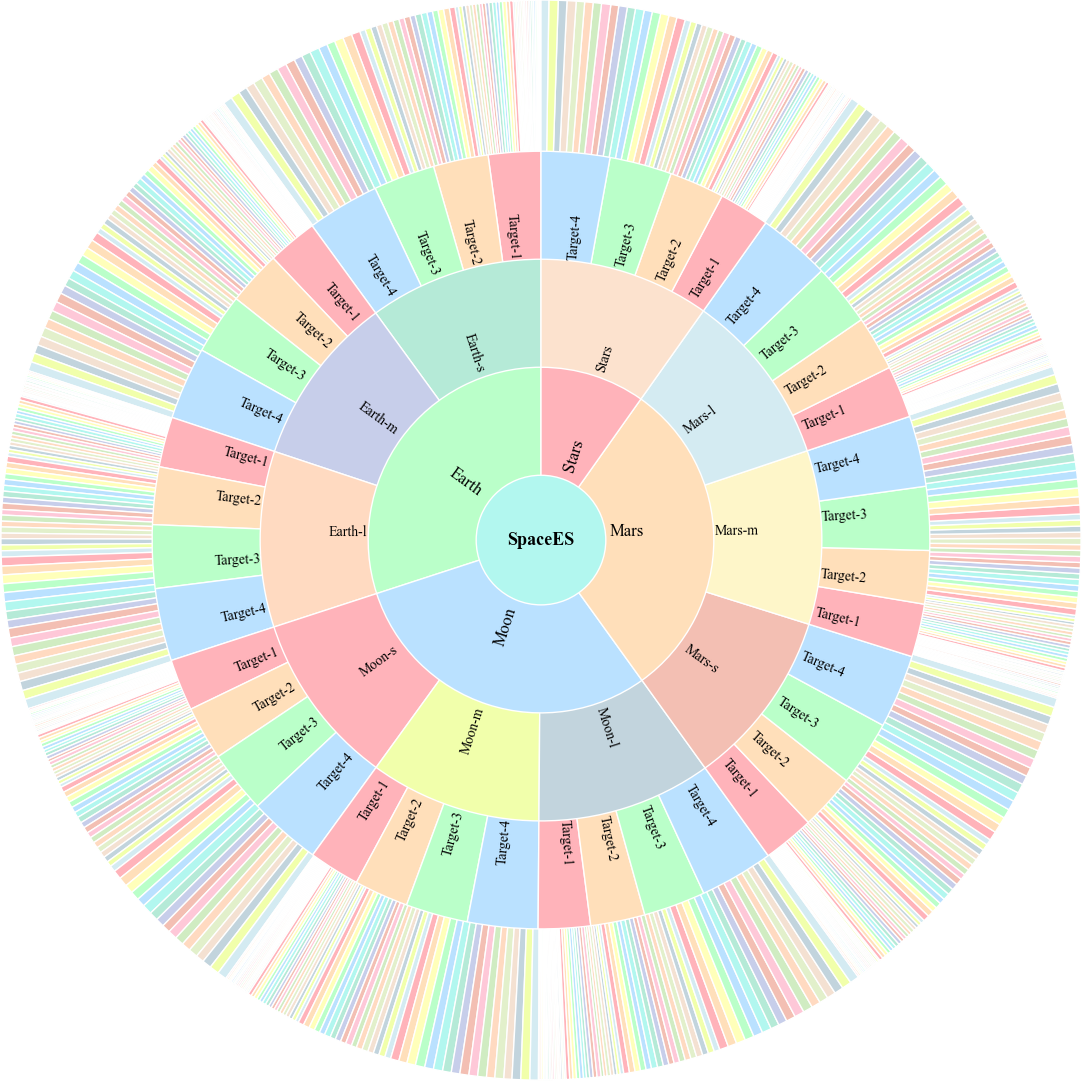}
    \caption{The distribution of categories and items in the SpaceES dataset, including data on different planets and cosmic background environments, various scales, varying numbers of targets, and multiple spacecraft model categories.} 
    \label{fig:S0}
\end{figure}

\begin{figure}[t]
    \centering
    \includegraphics[width=0.98\linewidth]{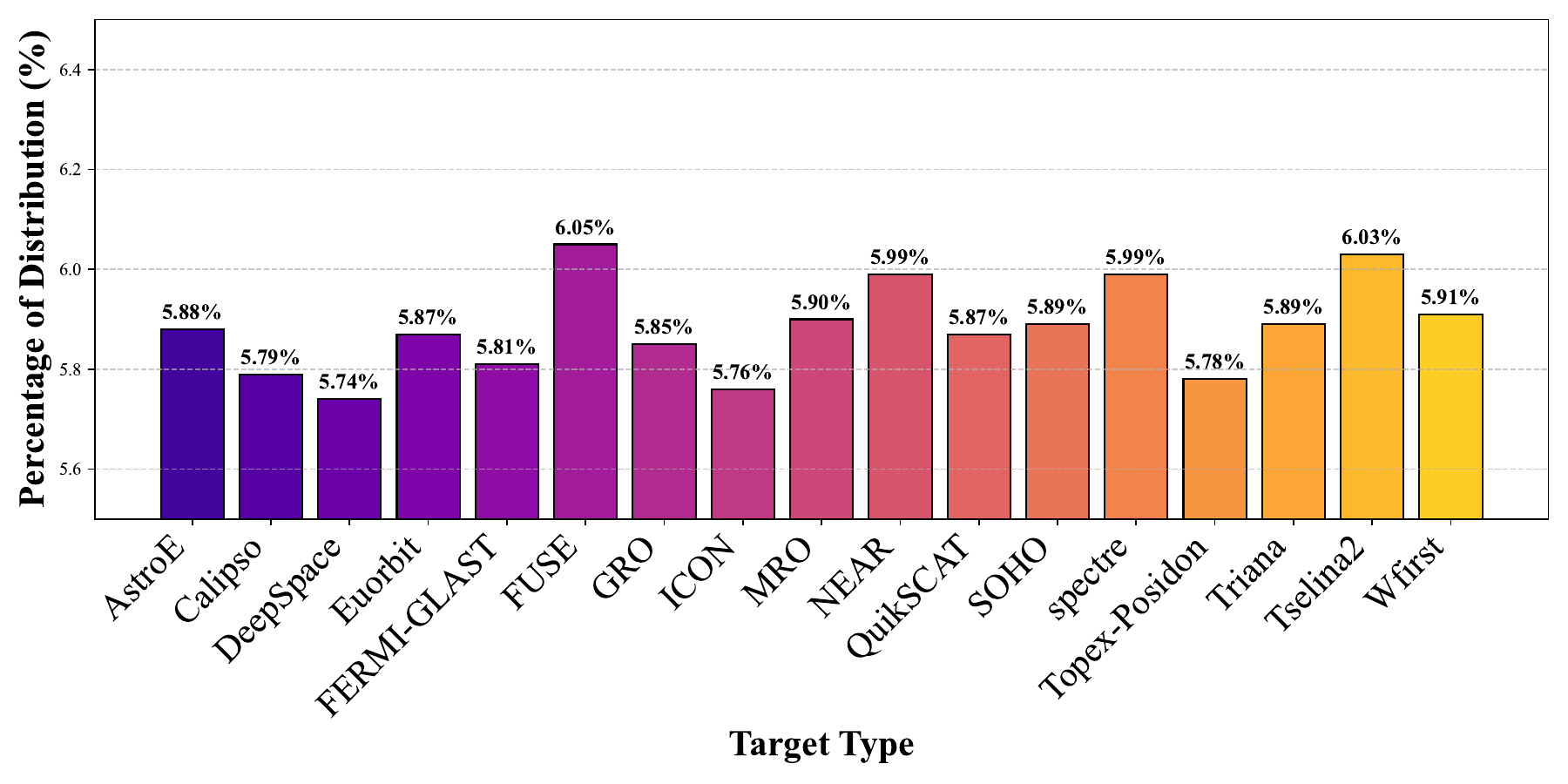}
    \caption{The distribution of spacecraft of various models in the entire dataset.} 
    \label{fig:S2}
\end{figure}

\begin{figure}[t]
    \centering
    \includegraphics[width=0.98\linewidth]{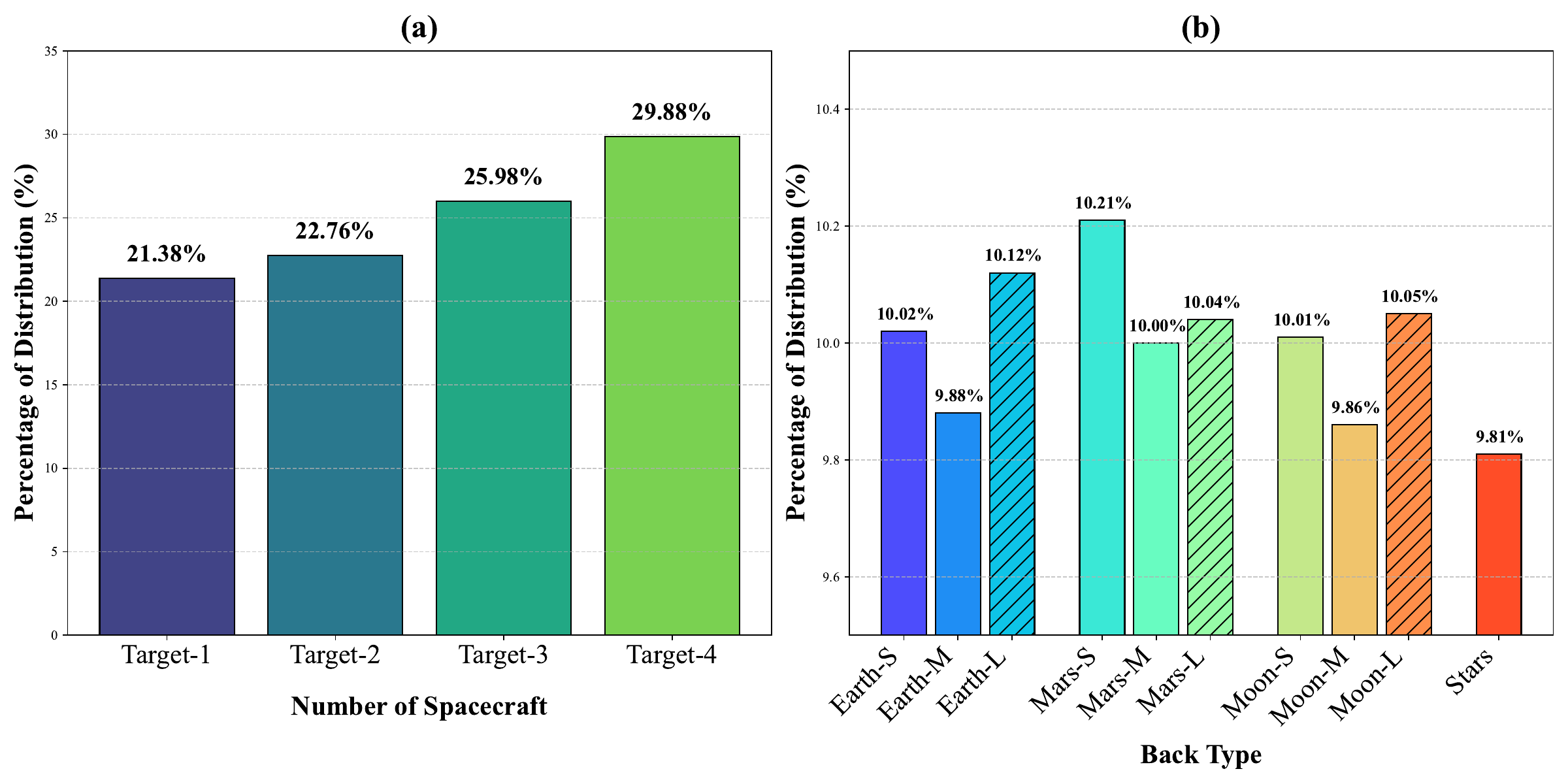}
    \caption{(a) is the category distribution of multi-spacecraft target imaging in the entire dataset. In the dataset, 78.62$\%$ of the samples consist of two or more on-orbit spacecraft targets. (b) is the distribution of imaging data at different scales across multiple spatial environments. Different spatial imaging environments and varying scales have a significant impact on visual tasks.} 
    \label{fig:S3}
\end{figure}

To validate our research, we conducted experiments using the spacecraft spatial target datasets, SpaceES, which was generated using our specially designed professional space data-driven engine, extracted following a balanced distribution, and its distribution pattern is shown in Figure \ref{fig:S0}. Based on simulations with various high-fidelity space backgrounds, the dataset defines five fundamental elements to simulate the spatial states of target spacecraft: number of images, feature types, operational attitudes, imaging sizes, and relative positions.The dataset spans scenarios involving 1 to 4 spacecraft, covering both single-target and multi-target configurations across 17 distinct target types [Figures \ref{fig:S2} and \ref{fig:S3}(a)]. For each target attitude, 2,592 frames were systematically sampled with randomized variations in imaging size and spatial positioning, ensuring comprehensive coverage of parameter space.

The dataset also takes into account three types of imaging interference factors in space exploration: blur, noise, and numerical disturbances. It constructs imaging data at different scales under the contexts of Earth, Moon, Mars, and starry background, with the distribution shown in Figure \ref{fig:S3}(b). Compared to existing space datasets, SpaceES offers extensive coverage, including various fly-around configurations such as space orbital constraints and natural fly-around configurations, double-ellipse spliced fly-around configurations, single droplet fly-around configurations, double droplet spliced fly-around configurations, and multi-pulse forced fly-around configurations. This comprehensive coverage enables it to realistically and reasonably meet the data training requirements of various space intelligent perception tasks.

The relative motion model of the sensing spacecraft with respect to the imaging target employs the Clohessy-Wiltshire (C-W) equations, with the analytical form of the state variation expressed as shown in Equation \eqref{cw}, where \([x(t),y(t),z(t),\dot{x}(t)),\dot{y}(t)),\dot{z}(t)]^T\) and \(\begin{bmatrix}x(0),y(0),z(0),\dot{x}(0)),\dot{y}(0)),\dot{z}(0)\end{bmatrix}^T\) represent the orbital state at time t and the initial state of the servicing spacecraft, respectively; \(n = \sqrt{\frac{\mu}{R_C^3}}\) denotes the orbital angular velocity of the sensing spacecraft; \(\mu\) is the Earth's gravitational constant; and \(R_c\) is the semi-major axis of the servicing spacecraft's orbit.

\begin{figure*}[t]
\begin{equation}
\label{cw}
    \begin{bmatrix}
        x(t) \\
        y(t) \\ 
        z(t) \\ 
        \dot{x}(t) \\ 
        \dot{y}(t) \\ 
        \dot{z}(t)
        \end{bmatrix}
    =
    \begin{bmatrix}
        4 - 3\cos nt & 0 & 0 & \frac{\sin nt}{n} & \frac{2 - 2\cos nt}{n} & 0 \\
        6(\sin nt - nt) & 1 & 0 & \frac{2\cos nt - 2}{n} & -3t + \frac{4\sin nt}{n} & 0 \\
        0 & 0 & \cos nt & 0 & 0 & \frac{\sin nt}{n} \\
        3n\sin nt & 0 & 0 & \cos nt & 2\sin nt & 0 \\
        6n(\cos nt - 1) & 0 & 0 & -2\sin nt & 4\cos nt - 3 & 0 \\
        0 & 0 & -n\sin nt & 0 & 0 & \cos nt
    \end{bmatrix}
    \begin{bmatrix}
        x(0) \\ 
        y(0) \\ 
        z(0) \\ 
        \dot{x}(0) \\ 
        \dot{y}(0) \\ 
        \dot{z}(0)
    \end{bmatrix}
\end{equation}
\end{figure*}

\subsection{Experiment Implementation}
Our method is implemented using PyTorch and trained on a single NVIDIA RTX 4090 GPU with 24 GB of memory. We utilize the AdamW optimizer with an initial learning rate of \(10^{-5}\). Unless otherwise specified, we use the Hiera-S variant of SAM2. Images are resized to \(1024 \times 1024\), and each training session involves 100,000 iterations of image data training.
\subsubsection{Comparative Experiments}
In the model comparison stage, we utilized MMSegmentation\cite{mmseg2020} as the foundational framework to implement multiple mainstream segmentation models and applied them to our constructed dataset, SpaceES. All models were trained using the same training strategy to ensure fair comparison. The test results are shown in Table \ref{taba}. From the experimental results, it can be observed that the proposed SpaceSeg model achieves optimal performance in both mIoU (\textbf{89.87$\%$}) and mAcc (\textbf{99.98$\%$}), surpassing mainstream segmentation methods. The parameter size is at a moderate level (55.34M), meeting the requirements of high-precision application scenarios.

\begin{table}[!htbp]
\centering
\caption{Targets Instance Segmentation Comparative Results}
\label{taba}
\begin{tabular}{lccc}
\toprule
Model & mIoU$\uparrow$ & mAcc$\uparrow$ & Params(M)$\downarrow$ \\
\midrule
ANNNet\cite{zhu2019asymmetricnonlocalneuralnetworks} & 62.06 & 72.75 & 43.86 \\
APCNet\cite{8954288} & 64.82 & 72.90 & 53.98 \\
CCNet\cite{Huang_2019_ICCV} & 65.29 & 75.03 & 47.46 \\
DeepLabv3+\cite{DBLP:journals/corr/abs-1802-02611} & 65.20 & 72.92 & 41.23 \\
DMNet\cite{9150886} & 63.83 & 70.99 & 50.81 \\
EncNet\cite{DBLP:journals/corr/abs-1803-08904} & 59.04 & 68.85 & 33.53 \\
FCN\cite{DBLP:journals/corr/LongSD14} & 66.37 & 75.77 & 47.13 \\
GCNet\cite{DBLP:journals/corr/abs-1904-11492} & 66.69 & 75.11 & 47.27 \\
HRNet\cite{DBLP:journals/corr/abs-1908-07919} & 44.73 & 55.00 & 9.64 \\
ISANet\cite{DBLP:journals/corr/abs-1907-12273} & 46.60 & 56.57 & 35.34 \\
K-Net\cite{DBLP:journals/corr/abs-2106-14855} & 81.34 & 88.99 & 60.34 \\
MAE\cite{DBLP:journals/corr/abs-2111-06377} & 77.48 & 85.84 & 162 \\
Mask2Former\cite{DBLP:journals/corr/abs-2112-01527} & 81.72 & 90.17 & - \\
NonLocal Net\cite{DBLP:journals/corr/abs-1711-07971} & 66.53 & 76.14 & 47.66 \\
OCRNet\cite{DBLP:journals/corr/abs-1909-11065} & 68.96 & 75.29 & - \\
PointRend\cite{DBLP:journals/corr/abs-1912-08193} & 78.04 & 86.09 & - \\
PSANet\cite{10.1007/978-3-030-01240-3_17} & 66.32 & 75.14 & 51.6 \\
PSPNet\cite{DBLP:journals/corr/ZhaoSQWJ16} & 63.76 & 70.03 & 46.61 \\
SAM2\cite{ravi2024sam2segmentimages} & 84.16 & - & - \\
SegFormer-B0\cite{DBLP:journals/corr/abs-2105-15203} & 79.87 & 89.14 & 3.72 \\
SegFormer-B1 & 82.20 & 90.51 & 13.68 \\
SegFormer-B2 & 82.96 & 91.50 & 24.73 \\
SegFormer-B3 & 83.43 & 91.08 & 44.6 \\
SegFormer-B4 & 83.60 & 90.99 & 61.37 \\
SegFormer-B5 & 82.21 & 90.89 & 81.97 \\
Segmenter\cite{DBLP:journals/corr/abs-2105-05633} & 51.90 & 63.27 & 102 \\
Semantic FPN\cite{DBLP:journals/corr/abs-1901-02446} & 44.58 & 55.29 & 28.5 \\
SETR\cite{zheng2021rethinkingsemanticsegmentationsequencetosequence} & 7.45 & 7.87 & 309 \\
Swin Transformer\cite{DBLP:journals/corr/abs-2103-14030} & 81.92 & 89.63 & 58.95 \\
Twins\cite{DBLP:journals/corr/abs-2104-13840} & 75.49 & 84.60 & 27.84 \\
UPerNet\cite{DBLP:journals/corr/abs-1807-10221} & 69.81 & 79.49 & 64.05 \\
Vision Transformer\cite{DBLP:journals/corr/abs-2010-11929} & 71.37 & 80.79 & 142 \\
\midrule
\textbf{SpaceSeg(Ours)} & \textbf{89.87} & \textbf{99.98} & 55.34 \\
\bottomrule
\end{tabular}
\end{table}

\subsubsection{Ablation Experiments}
To systematically evaluate the contributions of the proposed method and validate our design choices, we conducted functional ablation experiments. First, we investigated the effectiveness of our designed SDAT Framework by removing it. The experimental results showed a \textbf{0.78$\%$} drop in mIoU, which is a significant loss in high-precision scenarios. We also examined the impact of the MSHARD design. Reverting the decoder to the default SAM2 design resulted in a \textbf{5.71$\%$} decrease in segmentation accuracy, severely compromising the precision required for space object segmentation tasks. This fully demonstrates the critical role of MSHARD in improving segmentation accuracy.The comparative baseline for the experiments was established by training the SAM2 model with spatial selective dual-stream adaptive fine-tuning on the SpaceES spatial dataset. Table \ref{tabb} presents the quantitative results of different model variants.

\begin{table}[!htbp]
\centering
\caption{Design Validation through Functional Ablation Experiments}
\label{tabb}
\begin{tabular}{cccc}
\toprule
MSHARD & SDAT & mIoU$\uparrow$ & mAcc$\uparrow$ \\
\midrule
\faTimes & \faTimes & 84.16 & - \\
\faCheck & \faTimes & 89.09 & 99.97 \\
\faCheck & \faCheck & \textbf{89.87} & \textbf{99.98}\\
\bottomrule
\end{tabular}
\end{table}

\subsection{Hardware-in-the-Loop Simulations}

\begin{figure}[h]
    \centering
    \includegraphics[width=0.98\linewidth]{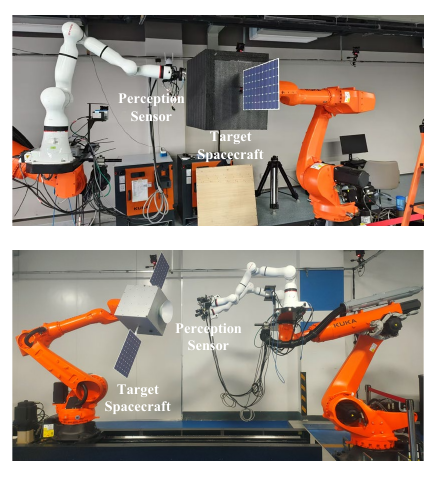}
    \caption{Simulated robotic arm devices for hardware-in-the-loop simulation of on-orbit spacecraft. Demonstration of on-orbit spacecraft operations in a ground test environment.} 
    \label{fig:hard}
\end{figure}

We conducted hardware-in-the-loop physical simulation experiments using KUKA robotic arms to simulate various dynamic characteristics and attitude changes of on-orbit spacecraft. This setup accurately reproduces the complex kinematic behaviors of on-orbit spacecraft, as illustrated in Figure \ref{fig:hard}. Our hardware-in-the-loop simulation demonstrates that the proposed segmentation model exhibits strong generalization capabilities, maintaining stable performance across various spacecraft types and orbital configurations. The model effectively meets the segmentation requirements for different spacecraft types and on-orbit spacecraft under varying attitudes, fully verifying its applicability to real-world space missions. In particular, it has significant application value in autonomous operation scenarios such as on-orbit servicing, space debris removal, and spacecraft docking, where reliable target segmentation capabilities are essential.

\subsection{Real-World Data Experiments}

\begin{figure}[h]
    \centering
    \includegraphics[width=0.98\linewidth]{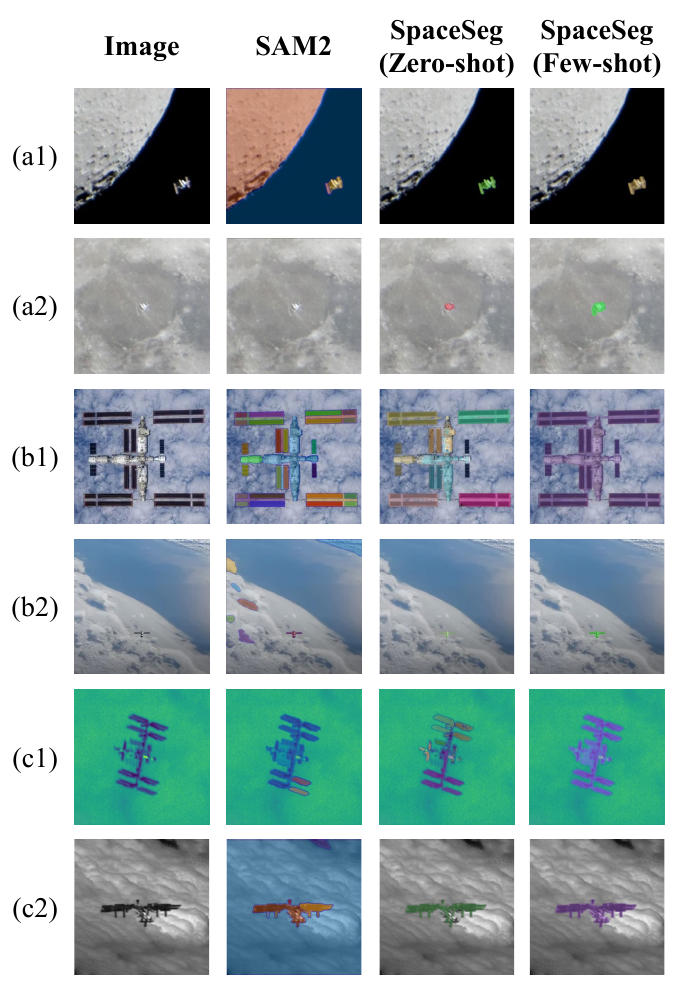}
    \caption{Real-world data experiments. Validation was performed using Zero-shot and Few-shot approaches. Figures a1 and a2 show the segmentation results under long-distance and celestial background imaging, respectively. Figures b1 and b2 represent the segmentation results for complex spacecraft and near-Earth orbit target tracking. Figures c1 and c2 illustrate the segmentation results under green band and near-infrared band imaging, respectively. The experimental results demonstrate that SpaceSeg significantly outperforms the baseline model SAM2 in handling various imaging scenarios. It exhibits superior segmentation accuracy and robustness when addressing diverse real spatial backgrounds and multi-scale imaging tasks. This fully validates the applicability and technical advantages of the SpaceSeg model in practical space missions.} 
    \label{fig:zf}
\end{figure}

We further validated the model using real-world satellite-to-satellite imagery data obtained from actual space missions, as shown in Figure \ref{fig:zf}. The data covers imaging under various complex interstellar observation conditions and across different bands (e.g., green band and near infrared band). The model delivers high-precision target segmentation results for real-world scene imaging, accurately delineating the contours of on-orbit spacecraft. The experimental results based on actual space imagery further confirm the model's robustness and practical value in real-world environments. Notably, the model maintains high-precision segmentation performance even under challenging conditions, such as complex lighting, varying observation distances, and realistic space backgrounds.

\section{CONCLUSION}
This paper proposes a novel multi-objective and multi-spatial background segmentation framework, SpaceSeg, specifically designed for on-orbit target segmentation tasks involving multiple spacecraft in a deep space environment. By introducing a multi-scale hierarchical attention refinement decoder, multi-spacecraft connected component analysis, a spatial domain adaptation transform framework and customized loss function design, SpaceSeg significantly improves model robustness and accuracy under the complex conditions of deep space. In addition, this paper constructs a multi-scale, multi-background dataset called SpaceES, which is specifically tailored for spacecraft intelligent perception. Comprehensive validation is conducted on both simulated and real spacecraft images, and experimental results demonstrate that the proposed method achieves state-of-the-art performance in both quantitative and qualitative evaluations. By combining the strengths of large-scale pretrained vision foundation models with innovative designs tailored for spacecraft perception tasks, SpaceSeg not only advances image segmentation technologies in deep space scenarios but also provides critical technical support for future space situational awareness missions.

\bibliographystyle{IEEEtran}
\bibliography{references}

\vfill

\end{document}